\pdfoutput=1

\documentclass[11pt]{article}

\usepackage[preprint]{acl}

\usepackage{times}
\usepackage{latexsym}
\usepackage[T1]{fontenc}
\usepackage[utf8]{inputenc}
\usepackage{microtype}
\usepackage{inconsolata}
\usepackage{graphicx}
\usepackage{booktabs}
\usepackage{amsmath}
\usepackage{tcolorbox}
\usepackage{amssymb}
\usepackage{multirow}
\usepackage{enumitem}
\usepackage{xspace}
\usepackage{colortbl}
\usepackage{arydshln}

\definecolor{myblue}{rgb}{0.0, 0.0, 0.8}
\definecolor{myorange}{rgb}{1.0, 0.647, 0.0}

\newcommand{\wo}{\emph{w/o}\xspace}

\title{CHOP: Mobile Operating Assistant with Constrained High-frequency Optimized Subtask Planning}

\author{
    Yuqi Zhou\textsuperscript{1},
    Shuai Wang\textsuperscript{2},
    Sunhao Dai\textsuperscript{1},
    Qinglin Jia\textsuperscript{2},
    Zhaocheng Du\textsuperscript{2},\\
    \textbf{Zhenhua Dong\textsuperscript{2}, Jun Xu\textsuperscript{1}\thanks{~~Corresponding author.}} \\
    \textsuperscript{1}Gaoling School of Artificial Intelligence, Renmin University of China \\
    \textsuperscript{2}Huawei Noah's Ark Lab\\
    \texttt{\{yuqizhou, sunhaodai, junxu\}@ruc.edu.cn}\\
}

\begin{document}
\maketitle
\begin{abstract}
The advancement of visual language models (VLMs) has enhanced mobile device operations, allowing simulated human-like actions to address user requirements. Current VLM-based mobile operating assistants can be structured into three levels: task, subtask, and action. The subtask level, linking high-level goals with low-level executable actions, is crucial for task completion but faces two challenges: \textbf{ineffective subtasks} that lower-level agent cannot execute and \textbf{inefficient subtasks} that fail to contribute to the completion of the higher-level task. These challenges stem from VLM’s lack of experience in decomposing subtasks within GUI scenarios in multi-agent architecture. To address these, we propose a new mobile assistant architecture with \textbf{c}onstrained \textbf{h}igh-frequency \textbf{o}ptimized \textbf{p}lanning (CHOP). Our approach overcomes the VLM's deficiency in GUI scenarios planning by using human-planned subtasks as the ``basis vector''. We evaluate our architecture in both English and Chinese contexts across 20 Apps, demonstrating significant improvements in both effectiveness and efficiency. Our dataset and code is available at \textcolor{blue}{\url{https://github.com/Yuqi-Zhou/CHOP}}
\end{abstract}

\section{Introduction}~\label{sec:introduction}
Mobile operating assistants~\cite{wang2024gui, zhang2024large, nguyen2024gui, hu2024agents} automate mobile App control by simulating human actions like clicking or typing. These assistants are widely used in recommendation~\cite{sun2022meta}, task automation~\cite{liu2024vision}, and user assistance~\cite{zhang2023appagent, wang2024mobile, zhu2024moba}. Early assistants, based on slot-filling and neural networks~\cite{sun2022meta, zhang2023you, zhu2023cam}, struggle with generalization. LLMs~\cite{openai2021chatgpt} improve this through multitask learning and cross-domain integration~\cite{brown2020language}, while VLMs~\cite{yang2024qwen2, openai2023gpt4} advance assistants by incorporating visual processing, making them the dominant approach in modern mobile environments~\cite{wang2024gui, zhang2024large, nguyen2024gui, hu2024agents}.

\begin{figure}[t]  
    \centering    
    \includegraphics[width=0.9\linewidth]{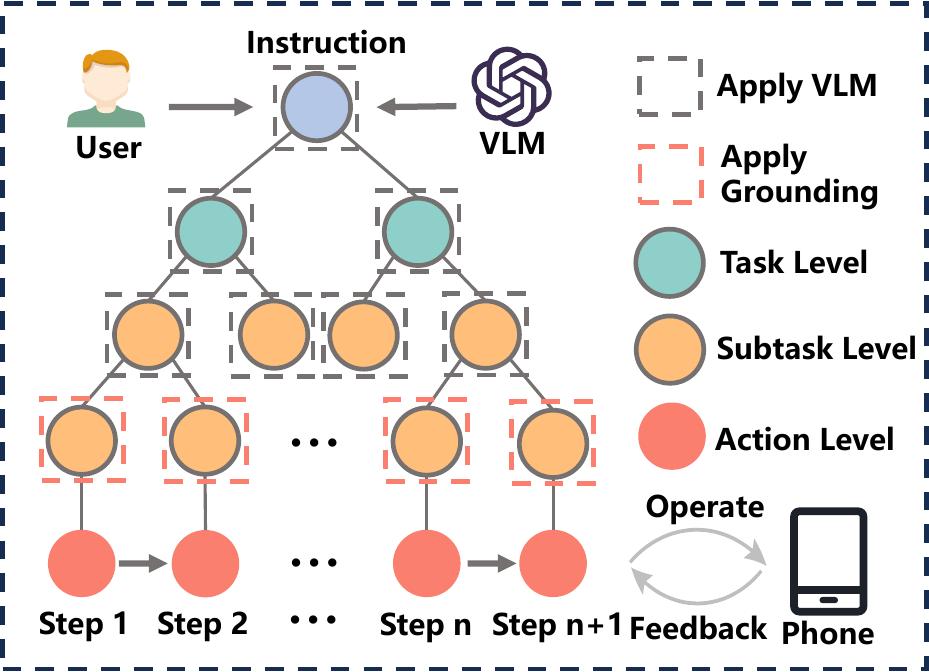}
    \caption{Execution flowchart for VLM-based assistant.}
    \label{fig:introduction}
    \vspace{-10pt}
\end{figure}

In mobile App operations, we structure VLM-based assistant architecture into three levels: task~\cite{chen2024octo}, subtask~\cite{zhu2024moba}, and action~\cite{lin2024showui, yang2024aria}, as shown in Figure~\ref{fig:introduction}. \textbf{A task} is a user directive within one App, typically consisting of multiple subtasks (e.g., ``Play Bob's songs''). \textbf{A subtask} is an independent instruction within a specific context, further decomposable into actions (e.g., ``Search Bob'' on the search interface). \textbf{An action} is the basic executable unit on the device (e.g., click). In this hierarchical architecture, a task is decomposed into subtasks, which are sequentially executed and translated into actions, enabling modules to cooperate in completing the task.

Although recent work in mobile assistants has attempted to improve subtask execution success by constraining the granularity of task decomposition~\cite{zhu2024moba}, subtask-level operations still face two main challenges: (1) \textbf{Ineffective subtasks}, where the subtask cannot be executed due to the VLM's lack of real-world knowledge~\cite{ahn2022can}. For instance, ``Go to Bob's office'' in response to ``Ask Bob to attend the meeting'' is unachievable, whereas ``Send Bob an email'' is more feasible. (2) \textbf{Inefficient subtasks}, where sequential actions unnecessarily delay task completion without contributing to progress. For example, ``Wait for Bob’s feedback'' stalls the task without advancing it. These challenges stem from VLM’s lack of experience in decomposing sub-tasks within GUI scenarios in multi-agent frameworks.

To address these challenges, we propose CHOP (\textbf{C}onstrained \textbf{H}igh-frequency \textbf{O}ptimized Subtask \textbf{P}lanning), a method that optimizes subtask planning by using basis subtasks as constraints during task decomposition. Specifically, in GUI scenarios, the same subtasks across different Apps share common operational logic, allowing users to quickly adapt to new Apps. This allows us to collect such subtasks and apply them to the task decomposition of the plan agent, meaning any task can be decomposed into a combination of ``basis subtasks'', inspired by ``basis vectors''. Meanwhile, we ensure the orthogonality of different basis subtasks by merging similar subtasks~\cite{wu2024atlas}. Furthermore, to better leverage the fixed-flow nature of basis subtasks, we provide documentation for each subtask to enhance effectiveness and allow the action agent to generate multiple steps in a single forward pass, thereby improving efficiency.

\begin{figure*}[t]  
    \centering    
    \includegraphics[width=1.0\linewidth]{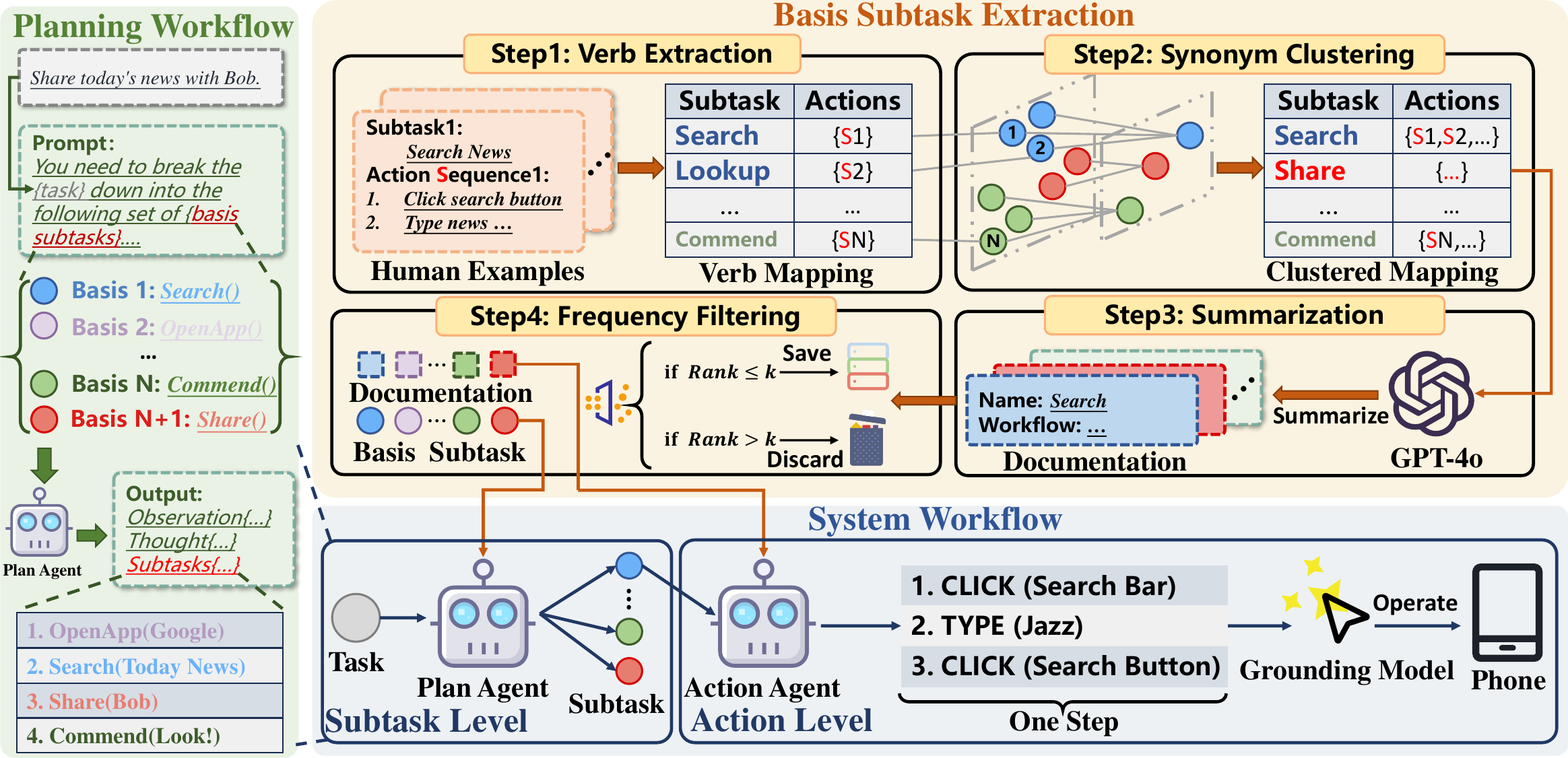}
    \caption{Illustration of the VLM-based GUI assistant framework with basis subtask extraction.}
    \label{fig:workflow}
    \vspace{-4pt}
\end{figure*}

We evaluate CHOP in both English and Chinese contexts. CHOP-En, the English dataset, is based on Mobile-Agent-V2~\cite{wang2024mobile}, covering 10 apps with three difficulty levels each. To extend this work to a broader linguistic context, we introduce CHOP-ZH, \textbf{the first Chinese dataset with user planning processes.} CHOP-ZH is created by hiring 10 annotators to complete 200 daily usage instructions across 10 apps, with annotators providing a plan and reasoning for each action. This allows us to evaluate the quality of the subtasks generated by the agent. We assess CHOP in terms of both effectiveness and efficiency, introducing new metrics to measure the inference cost of the action agent, grounding model, and overall architecture. Experimental results show that CHOP achieves state-of-the-art (SOTA) performance, outperforming mainstream VLM-based assistants.

Our summarized contributions are as follows: ($1$) We propose a new architecture, CHOP, which introduces ``basis subtasks'' for the first time and addresses the lack of planning capability in VLMs for GUI scenarios. ($2$) We construct the first Chinese dataset with user planning processes and introduce three new metrics for evaluating efficiency. ($3$) CHOP achieves SOTA performance on both English and Chinese datasets, with experimental results showing it generates higher-quality subtasks.

\section{Related Work}~\label{sec:related_work}
\textbf{GUI Agent.}
GUI agents have evolved from rule-based control to multimodal and reasoning-driven approaches. Early methods rely on predefined scripts but struggle in dynamic environments~\cite{li2017sugilite,li2019pumice}. Multimodal pre-trained models enabled end-to-end learning, integrating dialogue, screenshots, and operation history for better task execution~\cite{bai2021uibert,he2021actionbert,lispotlight,li2021screen2vec,wang2021screen2words,sun2022meta,zhang2023you}. In the era of VLMs, GUI agents incorporated complex reasoning and tool learning~\cite{qu2025tool,qu2024towards,quexploration}, using structured information in the view hierarchy to locate UI elements, thus improving efficiency and enabling deployment on devices~\cite{lee2024mobilegpt,zhang2024ufo,zhang2023appagent}. Image-only methods address cases without view hierarchy but remain challenged in dynamic settings~\cite{hong2024cogagent,wang2024mobile,zhu2024moba,zhang2024android}. Despite improving adaptability, VLM-based GUI agents still rely on VLMs that lack app-specific contextual knowledge. We address this gap by integrating structured human planning experience into the pipeline without requiring model fine-tuning.

\textbf{Multi-agent Application.}
LLMs possess strong comprehension and reasoning abilities, enabling LLM-based agents to autonomously execute tasks~\cite{wang2024survey,guo2024large}. Inspired by human collaboration, multi-agent frameworks are widely adopted, such as Smallville~\cite{park2023generative} and role-playing-based frameworks~\cite{li2023camel}. Recent advances include expert-agent coordination~\cite{chen2023agentverse}, meta-programming~\cite{hongmetagpt}, and multi-agent debating~\cite{chanchateval}. In GUI agents, multi-agent frameworks~\cite{wang2024mobile,zhu2024moba} often involve a plan agent for task planning, an action agent for interaction, and a grounding model that maps outputs to executable commands. However, these methods focus on introducing new modules while overlooking coordination among modules. Moreover, although Moba~\cite{zhu2024moba} also considers decomposing tasks multiple times to ensure the generated subtasks can be executed by the action agent, the issues of ineffective and inefficient subtasks we mentioned still persist. Instead, we propose constraining subtask-level outputs to improve executability by action-level agents and better facilitate task-level goals.

\section{Method}~\label{sec:method}
CHOP is an end-to-end pipeline that executes user instructions on real-world mobile devices, similar to~\cite{zhang2023appagent,wang2024mobile,zhu2024moba}. As shown in Figure~\ref{fig:workflow}, we present the CHOP and the extraction processes of its basis subtasks. \S~\ref{sec:problem_setup} first introduces the problem setup and environment construction. Then, \S~\ref{sec:base_subtask} outlines the extraction of basis subtasks used in task decomposition. Finally, \S~\ref{sec:framework} describes how CHOP integrates basis subtasks into its architecture, which consists of both the plan agent for task decomposition and the action agent for executing actions.

\subsection{Problem Setup}~\label{sec:problem_setup}
A mobile operating task consists of a screen $s$ and an instruction $q$ (e.g., ``Send an email to Bob''). Given a tuple $(s, q)$, a mobile operating assistant $f$ decides and performs a sequence of actions $\mathbf{a} = \{ a_1, a_2, \dots, a_t, \dots \}$ to interact with the Android environment $\mathcal{E}$ on the mobile device. This task execution is modeled as a sequential decision-making process. The formal definitions of the action and state spaces are as follows:

\begin{table}[t]
\centering
\resizebox{0.5\textwidth}{!}{%
\begin{tabular}{lll}
\toprule
Action Type & Attributes                                                                                                              & Description                  \\
\midrule
\texttt{CLICK}       & (x, y): Screen coordinates                                                                          & Click at an element          \\

\texttt{SCROLL}      & (direction): One of up, down, left, right       & Scroll the page       \\

\texttt{TYPE}        & (text): Text input                                                                                       & Type text                    \\
\hline
\texttt{BACK}        & -                                                                                                                       & Back to previous page \\

\texttt{EXIT}        & -                                                                                                                       & Task complete                \\

\texttt{WAIT}        & (time): Wait time in seconds                                                                                  & Stop for a while             \\
\bottomrule
\end{tabular}}
\caption{The supported action space for CHOP.}
\label{tab:action_space}
\vspace{-15pt}
\end{table}

\textbf{Action Space $A$}: We define an action as a function call~\cite{niu2024screenagent}. When the assistant outputs an action in the required format, it is parsed and executed by the environment. This includes various action types such as click, scroll, and type. Table~\ref{tab:action_space} provides a detailed list of action types and their corresponding attributes. \textbf{State Space $S$}: Since CHOP is an image-only architecture, it does not use textual information such as XML to assist decision-making. Instead, the state space is defined solely by the current screenshot $s_t$, which represents the environment at time step $t$.

At each time step $t$, the assistant selects an action $a_t$ based on the current state $s_t$ and the accumulated history $H_t=\{s_0,a_0,\dots,s_{t-1}, a_{t-1} \}$, as determined by the policy function: $a_t = f(s_t, H_t)$. The action $a_t$ leads to a state transition, where the Android environment $\mathcal{E}$ updates the state from $s_t$ to $s_{t+1}$ by the transition function $T$, reflecting the environmental changes resulting from the action: $s_{t+1} = T(s_t,a_t)$. At the same time, the history $H_t$ is updated to incorporate the most recent action $a_t$ and the previous state $s_{t-1}$, which results in: $H_{t+1} = \text{concat}(H_t, s_{t-1}, a_t)$.

In summary, the decision-making process begins with the initial state $S_0$, which represents the homepage of the mobile phone, and the initial history $H_0$, which is empty at the start. The assistant then proceeds by iterating through the policy $f$ and the transition function $T$, selecting an action at each time step $t$ and updating the state $s_t$ and history $H_t$. This continues until the action is \texttt{EXIT} or the maximum number of rounds is reached.

\subsection{Basis Subtask Extraction}~\label{sec:base_subtask}
Before introducing CHOP, we highlight two issues with subtask generation in the current multi-agent architecture: (1) \textbf{Ineffective subtasks}, where the plan agent generates unachievable subtasks due to the lack of real-world execution knowledge in VLMs~\cite{ahn2022can}. For example, ``Go to Bob's office'' in response to ``Ask Bob to attend the meeting'' is not executable, whereas ``Send email to Bob'' is more feasible. (2) \textbf{Inefficient subtasks,} where sequential execution increases task time without contributing to progress. For example, ``Wait for Bob’s feedback'' does not advance the task but prolongs execution.

To address these issues, ideal subtasks should meet two criteria: \textbf{High Effectiveness} – Executable by the action agent: The plan agent must generate subtasks that the action model can execute~\cite{ahn2022can}. \textbf{High Efficiency} – On the critical path: Any missing subtasks should lead to task failure, ensuring they are essential for task completion.

Inspired by human task planning~\cite{correa2023humans}, where individuals typically break down tasks based on familiar operations rather than methods that might seem optimal to others, we introduce basis subtasks—high-frequency subtasks commonly performed by humans. These subtasks enhance effectiveness (as they are familiar to humans due to their frequent use, making them easier to execute) and efficiency (since they are typically on the critical path of the task). 

Specifically, given the high cost of manually annotated data and the expensive fine-tuning of VLMs~\cite{lai2024autowebglm}, rather than training a new model, we focus on directly collecting these common subtasks from human-executed app commands to construct a ``basis subtask'' space. The collection process consists of four steps: Verb Extraction, Synonym Clustering, Summarization, and Frequency Filtering (Figure~\ref{fig:workflow}). \textbf{Clustering ensures that each basis subtask independently handles different task types, while filtering makes these ``basis subtasks'' easier to execute than others.} In summary, such subtasks can be seen as ``basis vectors''. Any task can be decomposed into a combination of independent basis subtasks, with their fixed nature enabling easier handling.

\textbf{Verb Extraction.} To capture subtasks, we use the AITZ dataset~\cite{zhang2024android}, a subset of AITW~\cite{rawlesandroidinthewild}, covering four Apps. Each entry in dataset contains an instruction and its step-by-step actions with the thought process. In AITW, raters annotate shorter sequences (at least $K \ge 3$ actions) as single-step demonstrations like “Add item to cart,” which are considered subtasks. Since verbs can represent actions, we use \textit{spaCy} for part-of-speech tagging, retaining only the verb to represent each instruction.

\textbf{Synonym Clustering.} Although verb extraction groups similar actions, synonyms with different expressions often serve the same function (e.g., “search news” vs. “lookup news”). Merging them reduces computational cost when generating subtasks~\cite{wu2024atlas}. To cluster words by semantic similarity, we use \textit{WordNet}\footnote{\url{https://github.com/argilla-io/spacy-wordnet}} to group them into synonym sets (synsets). Words are clustered based on shared synsets, reflecting their semantic similarity. After manual review, we retained verbs that represent meaningful actions and merged their corresponding action sequences.

\textbf{Summarization.}  In GUIs, consistent logic is applied across software to enhance user experience. For example, ``Search'' in browsers and email Apps follows similar steps: ``1. Click search box, 2. Enter content, 3. Click search button.'' Thus, action sequences within the same basis subtask should have similar representations. We standardize these sequences for downstream action agent to improve performance. Specifically, for each basis subtask, we use GPT-4 to summarize its corresponding action sequences with the prompt: ``Please summarize the following {{action sequence}} into a standardized process and specify boundary conditions.''

\textbf{Frequency Filtering.} Due to the performance degradation and increased inference time associated with longer input sequences, it is necessary to filter out certain basis subtasks. Since those basis subtasks that are more frequently used by humans in AITZ are likely to appear more often in the critical path, we rank them based on their frequency in the dataset and retain the top 10 most common basis subtasks. This filtering process ensures that the selected high-frequency basis subtasks are better able to generalize to unseen software. All the basis subtasks can be found in Table~\ref{tab:basis_subtasks} in the Appendix. An example of a basis subtask and its corresponding documentation is provided below:

\begin{tcolorbox}[title=A Basis Subtask with Documentation]
\small
\textbf{Basis subtask:} Search Item (parameter: search term) \\
\textbf{Standardized process:} 1. Click on the search bar located at the designated area of the screen.
2. Type in the content specified by the search term parameter.
3. If applicable, select a search suggestion from the dropdown list that appears after typing.
4. Press enter or click on the search button to execute the search. \\ 
\textbf{Boundary conditions:} 
1. If the search term is not found, check for spelling errors.
2. If selecting a suggestion, ensure it is the correct item before proceeding.
3. If navigating to a specific website, ensure the URL is entered correctly in the address bar.
\end{tcolorbox}

\subsection{CHOP: The Multi-Agent Architecture}~\label{sec:framework}
To guide the assistant $f$ in multi-step tasks, VLMs~\cite{openai2023gpt4,yang2024qwen2} are a strong candidate due to their visual understanding in mobile environments. However, applying VLMs to real-world screenshots with thousands of tokens is inefficient. Recent work~\cite{zhu2024moba} uses a two-stage architecture: decomposing tasks into subtasks and executing them, reducing sequence length, and improving accuracy~\cite{wang2024mobile}. However, without subtask constraints, ineffective and inefficient subtasks arise. To address these issues, we introduce basis subtasks during planning and limit outputs to predefined tasks, which incorporate human-designed heuristics to overcome VLM’s limitations in GUI scenarios. The process is described below.

\paragraph{The Plan Agent.} Given a user instruction $q$, the plan agent $f_\text{plan}$ decomposes it into a sequence of subtasks, each executable by the action agent:
$$
\{q_1,q_2,...,q_n\} = f_\text{plan}(q, Q_\text{basis}),
$$  
where $Q_\text{basis}$ is the set of predefined basis subtasks, and each $q_i$ must be selected from it. To enhance execution, the plan agent also generates the purpose and stopping condition for each subtask. If a necessary subtask is missing from $Q_\text{basis}$, a placeholder is used, prompting the model to define, structure, and refine new subtasks as needed. This ensures all generated subtasks are well-defined, actionable, and contribute effectively to task completion.

\paragraph{The Action Agent.} For each subtask $ q_i $, the action agent $ f_{\text{action}} $ determines the next executable action. At step $ t $, it generates an action $ a_{t+1} $ based on the user task $ q $, the current subtask $ q_i $, the execution documentation $ d_i $, the current screenshot $ s_t $, and the accumulated summary memories $ \mathbf{m} = \{m_1, \dots, m_{i-1}\} $. The selected action is then executed, updating the environment state:  
$$
a_{t+1} = f_{\text{action}}(q, q_i, d_i, s_t, \mathbf{m}),
$$  
$$
s_{t+1} = T(s_t, a_{t+1}).
$$  
To guide the execution of these actions, the agent generates observation, thought, and summarization. The summarization extracts key task-related details, such as weather information for the subtask ``Check today’s weather'', which is stored as memory $m_t$ for future tasks. Since VLMs output actions like \texttt{CLICK} without coordinates, we integrate Aria-UI~\cite{yang2024aria} to map these commands to precise locations (e.g., \texttt{CLICK}(Search Bar) $\rightarrow$ \texttt{CLICK}(200, 300)). To improve efficiency, $d_i$ provides standardized execution steps, and for basis subtasks with fixed workflows (e.g., ``Search item''), the agent generates the full action sequence in one step, minimizing latency and reducing the need for multiple action agent calls, which are a key source of computational bottleneck.

\section{Experiments}~\label{sec:experiments}
In this section, we evaluate the performance of CHOP by answering the following research questions: \textbf{RQ1:} Can the basis subtask improve overall task performance? \textbf{RQ2:} Can the basis subtask enhance the quality of task planning? \textbf{RQ3:} Can the basis subtask improve performance under certain conditions? RQ1 investigates whether adding the basis subtask constraint improves the execution of user instructions. RQ2 examines how the basis subtask affects the quality of subtasks generated by the plan agent. RQ3 analyzes the conditions under which the basis subtask demonstrates effectiveness in real-world, complex environments.

\subsection{Settings}

\paragraph{Test set.} We evaluate our method using two real-life scenario test datasets: \textbf{CHOP-En} and \textbf{CHOP-ZH}. The CHOP-En dataset consists of 30 English-language instructions, designed to test operating assistants in real-world mobile applications. It covers 10 widely used Apps in China, with tasks of varying difficulty levels: easy, medium, and difficult. The CHOP-ZH dataset consists of 200 Chinese instructions across 10 Apps, with 20 instructions per app. Annotators provided task plans alongside the instructions. This is the first real-life Chinese test set for mobile devices. In addition to instruction-action pairs, it enables a deeper evaluation of task decomposition. Due to resource constraints, we sample 3 instructions per app, as in CHOP-En. More details can be found in the Appendix~\ref{sec:dataset_details}.
\paragraph{Baselines.} To evaluate our method, we compare it with several baseline approaches, including the Human Baseline and agent-based automation methods. \textbf{Human Baseline} represents the ideal solution, reflecting the best performance achieved by a human. \textbf{AppAgent}~\cite{zhang2023appagent} employs an exploration-deployment framework where the agent learns app functions and uses these to plan and select actions. \textbf{Mobile Agent(v2)}~\cite{wang2024mobile} is a multi-agent system that integrates planning, decision-making, and reflection agents for mobile task automation, using screenshots and additional models like OCR and Qwen-VL-Plus. \textbf{Moba}~\cite{zhu2024moba} uses a two-level agent architecture (Global Agent and Local Agent), combining visual inputs and XML view hierarchy data for task planning and action execution. Detailed descriptions can be found in the Appendix~\ref{sec:baseline_details}.

\paragraph{Evaluation Metrics.} We evaluate the performance of assistants from two key aspects: \textbf{Effectiveness} and \textbf{Efficiency}. Effectiveness reflects the agent’s success in completing tasks, while Efficiency measures the speed and resource usage during task execution.
\textbf{Effectiveness:} We use two metrics: \textbf{Successful Rate (SR)} measures the proportion of tasks successfully completed within 20 actions. \textbf{Completion Rate (CR)}~\cite{zhu2024moba} evaluates the proportion of correct steps executed by the assistant, using human actions as the ground truth. 
\textbf{Efficiency:} To the best of our knowledge, we are the first to introduce the following three efficiency metrics for evaluating assistants:
\textbf{Mapping Efficiency (ME)} evaluates the efficiency of generating action sequences.
\textbf{Action Efficiency (AE)} measures the efficiency of executing actions.
\textbf{Average API Cost (AAC)} calculates the overall execution efficiency based on the number of API calls. Detailed formulas and calculations for these metrics are provided in the Appendix~\ref{sec:metrics_details}.

\paragraph{Experimental Setup.} All experiments are conducted using the \texttt{GPT-4o} model version to ensure a fair comparison. The maximum output length is set to $4096$, and the temperature during generation is set to $0.0$ to ensure reproducibility. The starting point for all instruction executions is set to the Homepage to ensure consistent evaluation. Due to the Moba method requiring additional tools to open the app, which are not available in our dataset, we use Aria-UI to handle app launching, as it ensures 100$\%$ accuracy. Unless specified, we will use CHOP-CH for the analysis experiments.

\begin{table*}[t]
\centering
\resizebox{1.0\textwidth}{!}{%
\begin{tabular}{llccccccccccccccc}
\toprule
\multicolumn{1}{c}{\multirow{3}{*}{\centering Language}} & \multicolumn{1}{c}{\multirow{3}{*}{\centering Model}} & \multicolumn{5}{c}{Easy} & \multicolumn{5}{c}{Medium} & \multicolumn{5}{c}{Hard} \\ 
\cmidrule(lr){3-7} \cmidrule(lr){8-12} \cmidrule(lr){13-17} & 
 & \multicolumn{2}{c}{Effectiveness} & \multicolumn{3}{c}{Efficiency} & \multicolumn{2}{c}{Effectiveness} & \multicolumn{3}{c}{Efficiency} & \multicolumn{2}{c}{Effectiveness} & \multicolumn{3}{c}{Efficiency} \\ 
\cmidrule(lr){3-4} \cmidrule(lr){5-7} \cmidrule(lr){8-9} \cmidrule(lr){10-12} \cmidrule(lr){13-14} \cmidrule(lr){15-17}
 &  & SR$\uparrow$ & CR$\uparrow$ & ME$\uparrow$ & AE$\uparrow$ & AAC$\downarrow$ & SR$\uparrow$ & CR$\uparrow$ & ME$\uparrow$ & AE$\uparrow$ & AAC$\downarrow$ & SR$\uparrow$ & CR$\uparrow$ & ME$\uparrow$ & AE$\uparrow$ & AAC$\downarrow$ \\ 
\midrule
\multirow{5}{*}{English} 
& Human           & 1.00 & 1.00 & 1.00 & 1.00 & - & 1.00 & 1.00 & 1.00 & 1.00 & - & 1.00 & 1.00 & 1.00 & 1.00 & - \\
\cmidrule(lr){2-17}
& AppAgent        & \underline{0.50}  & 0.62 & 0.84 & 0.84 & 1.19 & 0.40  & 0.64 & 0.80 & 0.80 & 1.25 & 0.10  & 0.22 & \underline{0.99} & \underline{0.99} & \underline{1.01} \\
& Mobile Agent(v2) & \underline{0.50}  & \underline{0.81} & 0.83 & 0.83 & 3.62 & \underline{0.50}  & \underline{0.73} & 0.82 & 0.82 & 3.65 & \underline{0.40}  & 0.41 & 0.68 & 0.68 & 4.42 \\
& Moba            & \underline{0.50}  & 0.69 & \underline{0.97} & \underline{0.97} & \underline{1.07} & 0.30  & 0.50 & \underline{0.99} & \textbf{0.99} & \underline{1.04} & 0.20  & \underline{0.46} & 0.98 & 0.98 & 1.05 \\
& Ours            & \textbf{0.80}  & \textbf{0.90} & \textbf{1.36} & \textbf{1.00} & \textbf{0.76} & \textbf{0.70}  & \textbf{0.89} & \textbf{1.20} & \underline{0.94} & \textbf{0.85} & \textbf{0.60}  & \textbf{0.59} & \textbf{1.10} & \textbf{1.00} & \textbf{0.93} \\ 
\midrule
\multirow{5}{*}{Chinese} 
& Human           & 1.00 & 1.00 & 1.00 & 1.00 & - & 1.00 & 1.00 & 1.00 & 1.00 & - & 1.00 & 1.00 & 1.00 & 1.00 & - \\
\cmidrule(lr){2-17}
& AppAgent    &    0.40          & 0.56          & 0.78          & 0.78          & 1.28          & {\underline{0.30}}    & 0.51          & {\underline{1.07}}    & 0.78          & 1.29          & {\underline{0.20}}    & 0.41          & {\underline{0.96}}    & \textbf{0.96} & {\underline{1.04}}    \\
& Mobile Agent(v2) &   {\underline{0.80}}    & {\underline{0.75}}    & 0.70          & 0.70          & 4.26          & 0.20          & 0.46          & 1.00          & {\underline{0.87}}    & 3.44          & \textbf{0.30} & {\underline{0.51}}    & 0.76          & 0.70          & 4.31          \\
& Moba            &  0.40          & 0.61          & {\underline{0.90}}    & {\underline{0.90}}    & {\underline{1.14}}    & {\underline{0.30}}    & {\underline{0.75}}    & 0.95          & 0.84          & {\underline{1.22}}    & 0.10          & 0.35          & 0.85          & 0.85          & 1.23          \\
& Ours            &  \textbf{1.00} & \textbf{1.00} & \textbf{1.30} & \textbf{0.95} & \textbf{0.79} & \textbf{0.80} & \textbf{0.95} & \textbf{1.10} & \textbf{0.95} & \textbf{0.93} & 0.10           & \textbf{0.59} & \textbf{1.09} & {\underline{0.93}}    & \textbf{0.95}      \\ 

\bottomrule
\end{tabular}
}
\caption{Performance evaluation of different GUI agents on English and Chinese tasks, categorized by difficulty. Metrics include effectiveness (\textbf{S}uccess \textbf{R}ate, \textbf{C}ompletion \textbf{R}ate) and efficiency (\textbf{M}apping \textbf{E}fficiency, \textbf{A}ction \textbf{E}fficiency, \textbf{A}verage \textbf{A}PI \textbf{C}ounts), with human as the baseline. Best results are bolded, and second-best are underlined.}~\label{tab:main_table}
\vspace{-10pt}
\end{table*}

\begin{table*}[t]
\centering
\resizebox{0.8\textwidth}{!}{%
\begin{tabular}{lcccccccccccc}
\toprule
\multirow{3}{*}{\centering Model} & \multicolumn{4}{c}{All (10 Apps)} & \multicolumn{4}{c}{In-domain (4 Apps)} & \multicolumn{4}{c}{Out-of-domain (6 Apps)} \\
\cmidrule(lr){2-5} \cmidrule(lr){6-9} \cmidrule(lr){10-13}
 & \multicolumn{2}{c}{Effectiveness} & \multicolumn{2}{c}{Efficiency} & \multicolumn{2}{c}{Effectiveness} & \multicolumn{2}{c}{Efficiency} & \multicolumn{2}{c}{Effectiveness} & \multicolumn{2}{c}{Efficiency} \\
\cmidrule(lr){2-3} \cmidrule(lr){4-5} \cmidrule(lr){6-7} \cmidrule(lr){8-9} \cmidrule(lr){10-11} \cmidrule(lr){12-13}
 & SR$\uparrow$ & CR$\uparrow$ & ME$\uparrow$ & AE$\uparrow$ & SR$\uparrow$ & CR$\uparrow$ & ME$\uparrow$ & AE$\uparrow$ & SR$\uparrow$ & CR$\uparrow$ & ME$\uparrow$ & AE$\uparrow$ \\
\midrule
CHOP            & \textbf{0.67} & \textbf{0.85} & \textbf{1.15} & \underline{0.91} & \textbf{0.75} & \textbf{0.92} & \textbf{1.31} & \underline{0.92} & \textbf{0.61} & \textbf{0.83} & \textbf{1.03} & \underline{0.80} \\
CHOP \wo $D_{\text{basis}}$      & \underline{0.47} & \underline{0.74} & \underline{1.00} & \textbf{1.00} & \underline{0.50} & \underline{0.73} & \textbf{1.00} & \textbf{1.00} & \underline{0.44} & \underline{0.76} & \underline{1.00} & \textbf{1.00} \\
CHOP \wo $Q_{\text{basis}}\&D_{\text{basis}}$   & 0.33 & 0.57 & 1.00 & \textbf{1.00} & 0.50 & 0.59 & 1.00 & \textbf{1.00} & 0.22 & 0.56 & 1.00 & \textbf{1.00} \\
\bottomrule
\end{tabular}
}
\caption{Ablation study on CHOP-ZH comparing the full method with two variants: one excluding the documentation $D_{\text{basis}}$ (CHOP \wo $D_{\text{basis}}$) and the other excluding both the basis subtask $Q_{\text{basis}}$ and $D_{\text{basis}}$ (CHOP \wo $Q_{\text{basis}} \& D_{\text{basis}}$). Experiments are conducted on three app sets: All (10 Apps), In-domain (4 Apps, where $Q_{\text{basis}}$ is collected), and Out-of-domain (6 Apps). The best results are bolded, second-best underlined.}
\label{tab:ablation_study}
\vspace{-5pt}
\end{table*}

\subsection{RQ1: Task Performance Improvement}

\paragraph{Main Results.} In RQ1, we investigate whether incorporating the basis subtask \( Q_{\text{basis}} \) and corresponding documentation \( D_{\text{basis}} \) into the plan agent's subtask generation improves the effectiveness and efficiency of CHOP. The main results are shown in Table~\ref{tab:main_table}, with human-executed trajectories serving as the ground truth. We compare CHOP with mainstream methods and draw the following conclusions:  

\textbf{(1) CHOP achieves the highest effectiveness:} CHOP outperforms other methods in \textbf{SR} and \textbf{CR} across most instruction sets. However, Mobile Agent(v2) outperforms CHOP on the Hard part of the Chinese dataset, likely due to CHOP's use of English documentation. \textbf{(2) CHOP demonstrates superior efficiency:} By generating multi-actions in one step for specific basis subtasks, CHOP achieves the best $\mathbf{ME}$ performance. It minimizes model calls with a single request to the plan agent. The high $\mathbf{AAC}$ confirms CHOP’s efficiency, using the fewest API calls and reducing resource consumption. \textbf{(3) Other methods show a trade-off between effectiveness and efficiency:} Mobile Agent(v2) offers comparable performance but requires at least three API calls per action, limiting practicality. AppAgent and Moba, though less efficient, perform well with good resource utilization.  

\paragraph{Ablation Study.} We draw two key conclusions from our experiments in Table~\ref{tab:ablation_study} on removing documentation and the basis subtask constraint during subtask generation.

\textbf{(1) Removing documentation and the basis subtask both reduce performance, highlighting the importance of these components.} Specifically, experiments show that CHOP’s performance decreases when documentation is excluded, and performance worsens further without the basis subtask. Additionally, CHOP’s \textbf{AE} score drops, likely due to the variants adopting simpler behaviors (e.g., searching for contacts directly instead of clicking avatars), requiring fewer actions. \textbf{(2) The basis subtask improves CHOP’s performance even on out-of-domain Apps, demonstrating its generalizability.} Although basis subtasks are collected from AITW (which includes four app types), experiments on both in-domain (same app types) and out-of-domain datasets show that the basis subtask benefits performance across both. This supports the idea that similar subtasks across Apps share common operational logic. Furthermore, compared to AppAgent which collects whole-app documentation, our approach reduces size to the subtask level, improving generalization and data efficiency.

\subsection{RQ2: Task Planning Improvement}
\paragraph{Subtask Evaluation} Unlike previous experiments that evaluated the performance of the entire architecture, we now focus on the quality of subtasks. Our evaluation examines two aspects: 

\textbf{(1) Matching Metrics:} In this study, we use two widely used metrics, BLEU~\cite{papineni2002bleu} and ROUGE-L~\cite{lin2004rouge}, to measure the similarity between two texts, with the subtasks annotated by labelers in CHOP-CH serving as the golden reference. A higher score indicates greater similarity. \textbf{(2) LLM as Evaluator}: Leveraging the strong performance of LLMs in text quality assessment~\cite{zhengjudging}, we use an LLM to evaluate the subtasks generated by the plan agent, both before and after incorporating the basis subtask. The evaluation focuses on three criteria: completeness (whether the subtasks can achieve the task’s goal when executed), efficiency (avoiding irrelevant subtasks), and effectiveness (whether the subtasks can be executed by the action agent). To mitigate token and position bias~\cite{dai2024bias}, we randomly shuffle the comparison objects prior to evaluation and calculate the winning proportions.

\begin{figure}[t]  
    \centering    
    \includegraphics[width=0.9\linewidth]{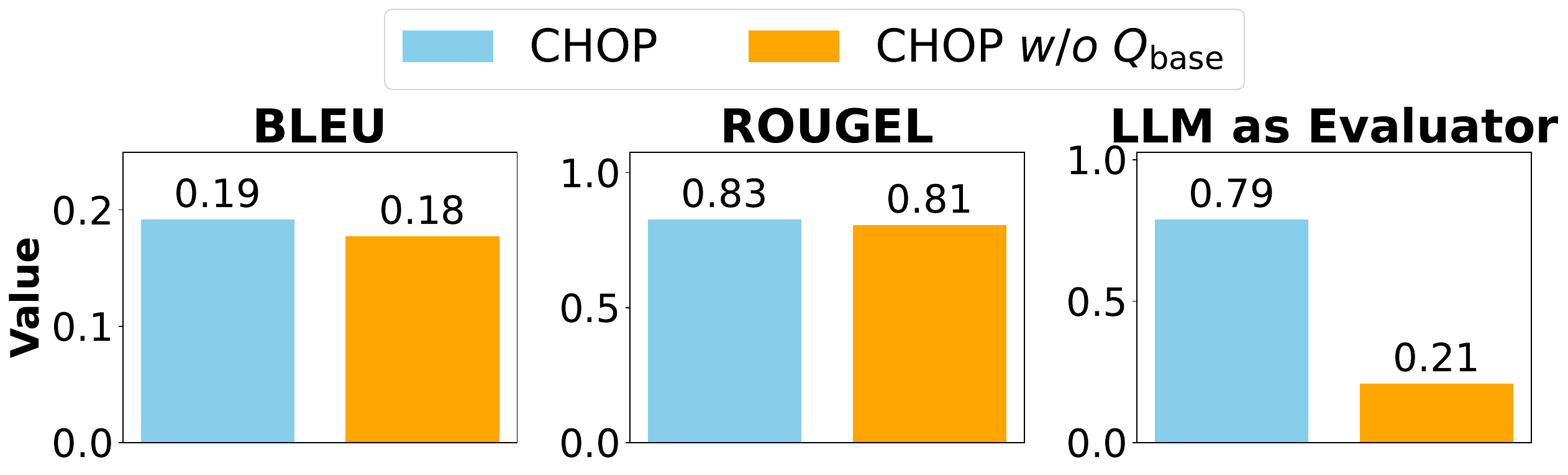}
    \caption{Subtask quality comparison with and without basis subtask on matching and LLM-based evaluation.}
    \label{fig:subtask_quality}  
    \vspace{-10pt}
\end{figure}

The detailed results are presented in Figure~\ref{fig:subtask_quality}. As shown, whether evaluated using token-level matching metrics or the LLM-based evaluation, the scores of subtasks generated after adding basis subtask constraints outperform the previous ones. This demonstrates that the basis subtask enhances the quality of the generated subtasks.

\paragraph{Case Study.} The plan agent is not only tasked with generating basis subtasks but also has the flexibility to create custom subtasks when the basis subtask is unavailable. As demonstrated in Appendix~\ref{sec:subtask_case}, we present two examples showing the task and its corresponding subtasks. These examples highlight that, in addition to effectively selecting basis subtasks, our method CHOP can also generate high-quality custom subtasks that effectively complement the basis subtasks. In addition, we also demonstrate with two examples that adding the constraint of basis subtasks can address the issues of ineffective and inefficient subtasks.

\subsection{RQ3: Conditions for Improvement}
\begin{figure}[t]  
    \centering    
    \includegraphics[width=0.7\linewidth]{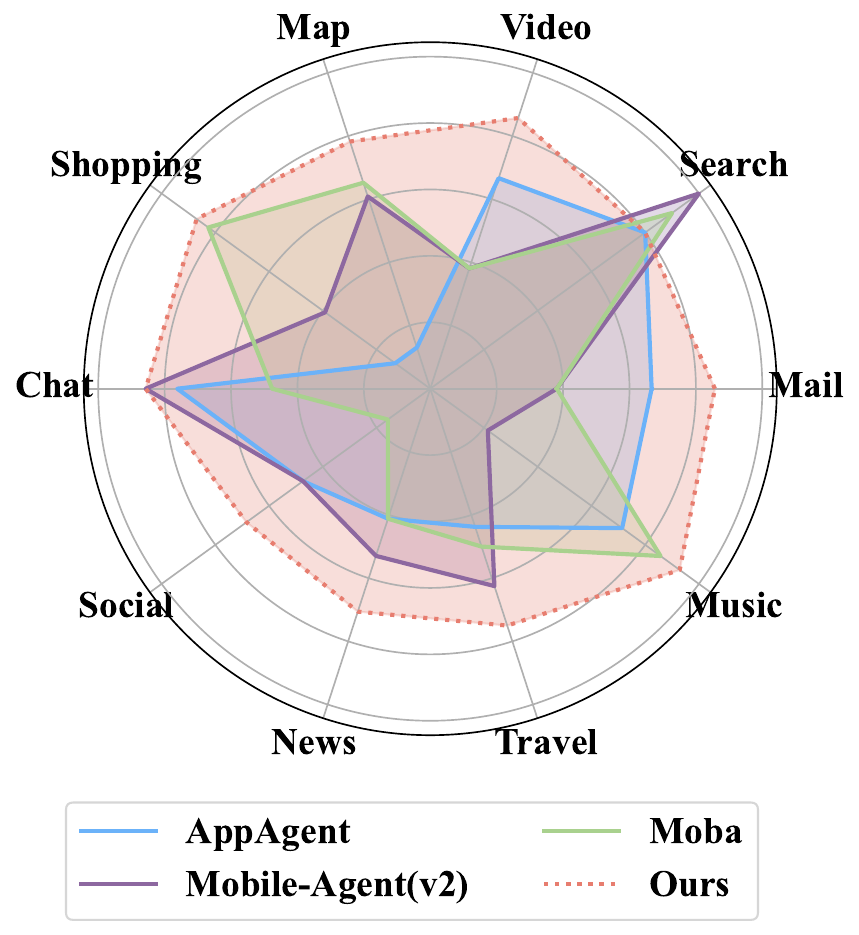}
    \caption{Performances of CHOP with other methods.}
    \label{fig:app_types}
    \vspace{-15pt}
\end{figure}

\paragraph{Improvement on Various App.}
\textbf{RQ3} analyzes which tasks benefit most from the basis subtask. We first calculate the \textbf{CR} metric for all methods across 10 different application categories. As shown in Figure~\ref{fig:app_types}, our method consistently achieves a high \textbf{CR} across various applications. In contrast, other methods like AppAgent struggle with app types such as Shopping and Map due to XML parsing issues, while our vision-based method bypasses this problem.

\begin{figure}[t]  
    \centering    
    \includegraphics[width=0.7\linewidth]{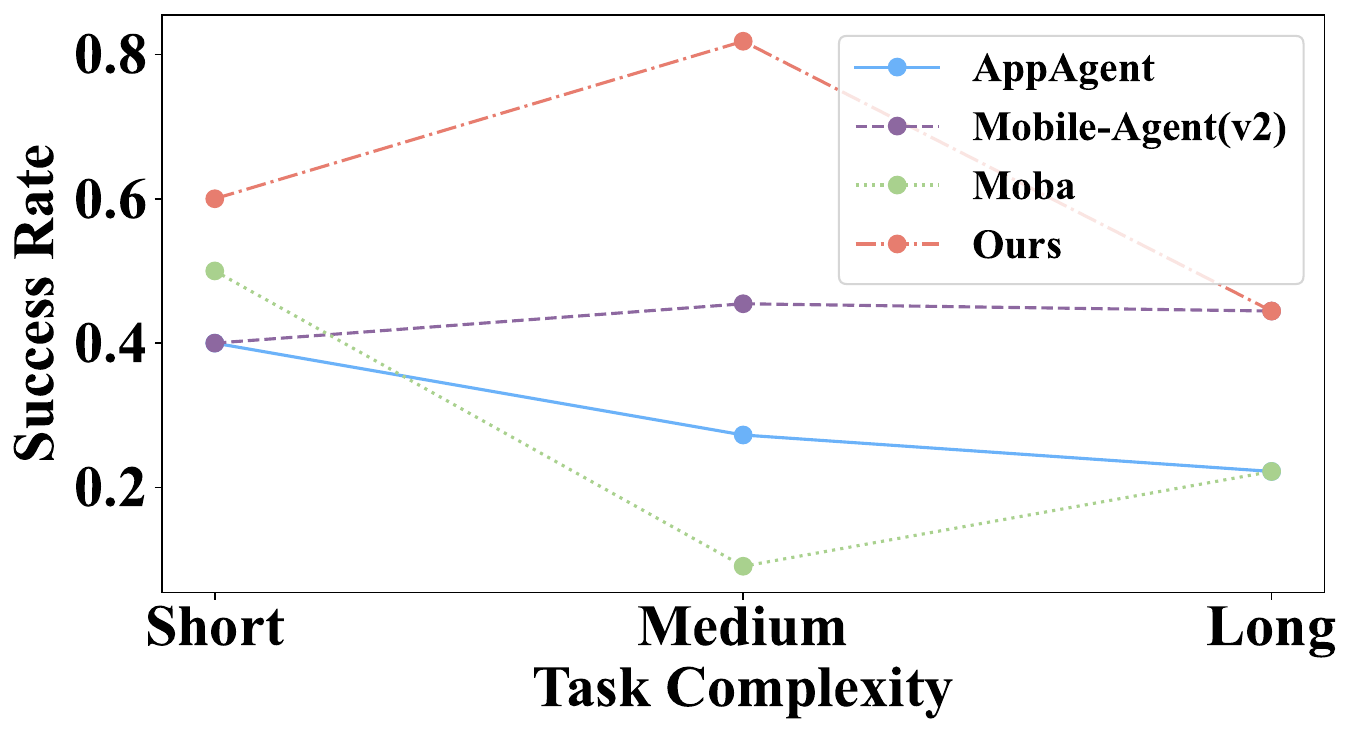}
    \caption{\textbf{SR} of different methods across tasks of varying complexities, where complexity is defined by task length, with segments based on consecutive echo points.}
    \label{fig:length_acc}  
\end{figure}

\paragraph{Improvement on Complex Instruction.}
We also measure \textbf{SR} on instructions of varying complexity, defined by step count. As shown in Figure~\ref{fig:length_acc}, we group instructions into three length segments. The results show that our method performs particularly well with short and medium-length instructions, with the largest improvement seen in medium-length tasks. However, the improvement is smaller for both short and long instructions. For short instructions, the bottleneck seems to lie outside task planning, likely in visual capabilities. For long instructions, the challenge is the higher requirement for successful subtask decomposition, but our method still outperforms others.

\paragraph{Error on Different Types.}
As shown in Table~\ref{tab:error_analysis}, we analyze failure reasons for various methods following the settings in~\cite{lai2024autowebglm}. Both AppAgent and Moba depend on XML files, so XML parsing errors lead to failures, while text-based output parsing errors also contribute. We categorize these as ``XML/Model Output Parse Error.'' AppAgent is most affected by XML parsing, highlighting the need for image-only solutions. Mobile-Agent(v2) and Moba show high ``Misinterpretation of Task Context'' rates, pointing to planning-level issues. In contrast, our approach has a low rate of this error, indicating that the basis subtask improves planning.

\begin{table}[t]
\resizebox{0.48\textwidth}{!}{%
\begin{tabular}{lcccc}
\toprule
Error Type                        & AppAgent & Mobile-Agent(v2) & Moba & Ours  \\
\midrule
Hallucinations                    & 4.8\%  & 5.9\%  & 9.1\%  & 0.0\%  \\
Poor Graphical Recognition        & 9.5\%  & 29.4\% & 9.1\%  & 54.6\% \\
Misinterpretation of Task Context & 23.8\% & 47.1\% & 63.6\% & 45.5\% \\
Exceeds Max Iterations            & 4.8\%  & 17.7\% & 4.6\%  & 0.0\%  \\
XML/Model Output Parse Error      & 57.1\% & 0.0\%  & 13.6\% & 0.0\%  \\
\bottomrule
\end{tabular}
}
\caption{Error distribution in mobile operating assistant.}~\label{tab:error_analysis}
\vspace{-20pt}
\end{table}

\paragraph{Case Study.} Finally, we demonstrate that our method enables agents to follow a more structured execution pattern, reducing errors and improving efficiency by generating multi-step actions in a single call. This leads to smoother task execution and faster completion times. A detailed explanation and figures can be found in Appendix~\ref{sec:case_study}.

\section{Conclusion}~\label{sec:conclusion}
We present CHOP, a mobile operating assistant that enhances task execution by leveraging basis subtasks extracted from high-frequency human-executed sequences. CHOP identifies these basis subtasks through four key steps: verb extraction, synonym clustering, summarization, and frequency filtering. By integrating basis subtasks into the planning process, CHOP ensures that generated subtasks are both executable and aligned with key task pathways, leading to improved task effectiveness and efficiency. Experimental results on English and Chinese datasets demonstrate significant gains in execution quality over existing methods, highlighting CHOP as a robust solution.

\section*{Limitations}~\label{sec:limitation}
We believe the proposed CHOP method represents a significant step forward in advancing GUI agent research in the LLM era. However, several limitations remain that should be addressed in future work. First, the current evaluation process relies on manual assessments, which results in a relatively small dataset. Future research should aim to develop an automated evaluation pipeline to handle large-scale data and provide more stable and reproducible results. Second, our work currently focuses on the issues between the planning agent and the action agent in a multi-agent architecture, without exploring the potential challenges between the action agent and the grounding model. Future efforts should investigate how to better enable the action agent to effectively utilize the grounding model. Finally, the current architecture enhances VLM’s planning capabilities in GUI scenarios through prompts, as searching for planning data is computationally expensive. However, fine-tuning directly on data offers a more reliable approach. Future research should explore the use of synthetic data for fine-tuning to further strengthen VLM’s planning capabilities.

\bibliography{custom}

\appendix

\clearpage
\newpage
\section{Test Set Details}~\label{sec:dataset_details}
To conduct an in-depth comparison of the ability of our method and other assistants to handle complex user instructions and task execution efficiency on mobile devices, we evaluate them on two real-life scenario test datasets, namely, CHOP-En and CHOP-ZH. 

The CHOP-En dataset consists of 30 instructions used to assess the performance of assistants in real-world mobile applications with a diverse set of English tasks. This dataset is collected following the setup of the dataset used in Mobile Agent(v2)~\cite{wang2024mobile}, where 10 widely used applications in China are selected, covering various everyday scenarios. For each application, three tasks of different levels of difficulty were included: easy, medium, and difficult. The easy-level instructions explicitly specify the app to be used and typically require fewer than five steps to complete. Medium-level instructions necessitate more actions to be executed, while difficult-level instructions are presented in natural language without specifying the app to be used.

The CHOP-ZH dataset consists of 200 human-curated and annotated Chinese instructions. The dataset is constructed by selecting 10 applications that cover a broad range of daily usage scenarios. For each application, annotators who are in-house data labelers first provide 20 instructions based on daily tasks and execute them on mobile phones. Before execution, annotators are asked to create a subtask plan for each task and describe their thought process before performing each action. Additionally, we anonymized all the data by replacing all personal information with placeholders. Compared to similar English task sets~\cite{zhang2023appagent,wang2024mobile}, the CHOP-ZH dataset is the first real-life \textbf{Chinese} test set designed for mobile devices. Additionally, while these datasets only provide instructions and corresponding actions for each step, the CHOP-ZH dataset offers a comprehensive task plan. This allows us not only to assess the overall performance of the architecture based on task execution but also to evaluate the plan agent's ability to decompose tasks, providing a more targeted evaluation. Due to the high cost of GPT-4o, we sample 3 instructions per app and assign them difficulty levels (easy, medium, hard) as in CHOP-En. The test instructions and CHOP-ZH details are in Table~\ref{tab:dataset_details}.

\begin{table}[t]
\resizebox{0.45\textwidth}{!}{%
\begin{tabular}{lccc}
\toprule
Dataset Name  & CHOP-En  & CHOP-ZH (Sampled) & CHOP-ZH (Full) \\
\midrule
\#Instructions & 30            & 30                    & 200                \\
\#Task Steps   & 5.57          & 5.53                  &                     \\
Language & English & Chinese & Chinese \\
Screen Image   & $\times$      & \checkmark            & \checkmark         \\
Plan Thought   & $\times$      & \checkmark            & \checkmark         \\
Action Thought & $\times$      & \checkmark            & \checkmark         \\
\bottomrule
\end{tabular}
}
\caption{Dataset details, including instruction count, task steps, and availability of supporting data.}~\label{tab:dataset_details}
\vspace{-20pt}
\end{table}

\begin{table*}[t]
\centering
\resizebox{1.0\textwidth}{!}{%
\begin{tabular}{p{0.4\textwidth}|p{0.55\textwidth}}
\toprule
\textbf{Task} & \textbf{Subtasks} \\
\midrule
\multirow{5}{0.4\textwidth}{Search for videos about Stephen Curry on Bilibili and open 'Comments' to comment 'Oh, chef, your basketball spirit has always inspired me'} 
& 1. Find App (Bilibili) \\
& 2. Search Item (Stephen Curry videos) \\
& \textcolor{red}{3. Open Video (Stephen Curry)} \\
& 4. Access Comments (Stephen Curry video) \\
& 5. Post Comment ('Oh, chef, your basketball spirit has always inspired me.') \\
\hline
\multirow{5}{0.4\textwidth}{Open the Calendar and look at today's date, then go to Notepad and create a new note to write 'Today is {[}today's date{]}'} 
& 1. Find App (Calendar) \\
& \textcolor{red}{2. Check Date (today's date)} \\
& 3. Back Home \\
& 4. Find App (Notepad) \\
& \textcolor{red}{5. Create New Note (Today is {[}today’s date{]})} \\
\bottomrule
\end{tabular}
}
\vspace{-5pt}
\caption{Two task examples with corresponding subtasks, with custom subtasks in red.}
\vspace{-5pt}
\label{tab:custome_subtask_example}
\end{table*}

\begin{table*}[t]
\centering
\resizebox{1.0\textwidth}{!}{%
\begin{tabular}{p{0.4\textwidth}|p{0.55\textwidth}}
\toprule
\textbf{Task} & \textbf{Subtasks} \\
\midrule
\multirow{8}{0.4\textwidth}{Share the latest video from Bilibili content creator Johnny with Bob on WeChat.} 
& 1. Open the Bilibili app or website. \\
& 2. \textcolor{myblue}{Find the latest video from content creator Johnny.} \\
& 3. Click the share button and select WeChat. \\
& 4. In the sharing interface, choose the contact Bob and send the video link. \\
\cdashline{2-2}
& 1. Find App (Bilibili). \\
& 2. Search Item (Johnny). \\
& 3. Open Section (Johnny). \\
& 4. Share Content (WeChat, Bob). \\
\hline
\multirow{6}{0.4\textwidth}{Could you please check my search history on Baidu?} 
& 1. Open the Baidu browser or Baidu app. \\
& 2. \textcolor{myorange}{Log in to your Baidu account (if not already logged in).} \\
& 3. Access the history option and open it to view your Baidu search history. \\
\cdashline{2-2}
& 1. Find App (Baidu) \\
& 2. Open Section (search history) \\
& 3. Check Notifications (search history) \\
\hline
\end{tabular}
}
\vspace{-5pt}
\caption{Task examples with corresponding subtasks, without the basis subtask restriction. Ineffective subtasks are in blue, and inefficiency is in orange.}
\label{tab:case_study}
\vspace{-10pt}
\end{table*}

\begin{table*}[t]
\centering
\resizebox{1.0\textwidth}{!}{%
\begin{tabular}{p{0.3\textwidth}|p{0.7\textwidth}}
\toprule
\textbf{Basis Subtask} & \textbf{Explanation} \\
\midrule
Search Item (parameter) & Click on the search bar, type in the item name, and press enter. The parameter is the name of the item you want to search for. This action can be performed on any website with a search functionality. Output format is ``Search Item (XXX)''. \\
\hline
Send Text Message (parameter) & This action involves typing a specific message into a designated text input area. The parameter is the content of the message to be sent. Output format is ``Send Text Message (XXX)''. \\
\hline
Open Section (parameter) & Find and enter the specified section or feature in the application. The parameter is the name of the section, such as ``Hot List'', ``Messages'', ``Settings'', etc. \\
\hline
View Content (parameter) & View the specified content in the application. The parameter describes the content to be viewed, such as ``Latest News'', ``Posts'', etc. \\
\hline
Interact (parameter1, parameter2) & Interact with the content in the application, such as ``Like'' or ``Comment''. The parameter1 is the content to interact with, such as ``Video'' or ``Song''. The parameter2 is the action of interaction, such as ``Like content'', ``Post a comment'', etc. \\
\hline
Manage Collections (parameter1, parameter2) & Manage personal collections or shopping carts, etc. The parameter1 includes actions such as ``Add to Favorites'', ``Delete'', and parameter2 includes items such as ``Product'', ``Video'', etc. \\
\hline
Share Content (parameter1, parameter2) & Share content from the application to other platforms or users. The parameter1 includes the sharing platform and parameter2 includes the recipient, such as ``WeChat'', ``Lucky''. \\
\hline
Check Notifications (parameter) & View notifications or messages in the application. The parameter is the section of the app, such as ``System Notifications'', ``Private Messages'', etc. \\
\hline
Modify Settings (parameter1, parameter2) & Modify the settings in the application. The parameter1 includes the setting item and parameter2 includes its changes, such as ``Theme Skin'', ``Notification Method'', etc. \\
\hline
Create or Edit Entry (parameter1, parameter2) & Create or edit entries in the application. The parameters include the entry type and name, such as ``Playlist'', ``Contact'', etc. \\
\bottomrule
\end{tabular}
}
\caption{Description of various basis subtasks and their explanations.}
\label{tab:basis_subtasks}
\end{table*}

\begin{figure*}[ht]  
    \centering    
    \includegraphics[width=1.0\linewidth]{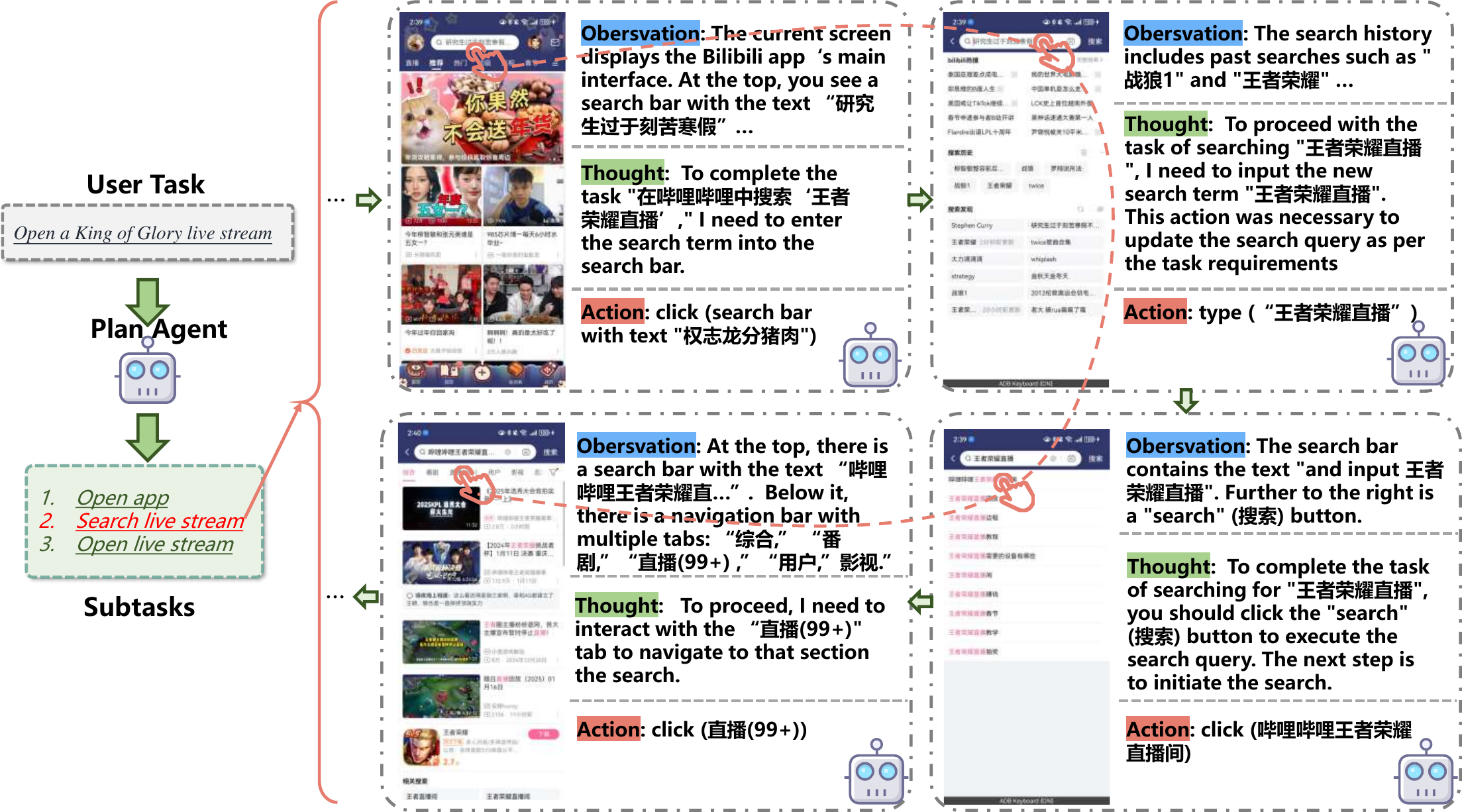}
    \caption{Subtask: Search live stream.}
    \label{fig:case_study}  
\end{figure*}

\section{Baseline Details}~\label{sec:baseline_details}
To provide a comprehensive evaluation, we also implement several baseline methods for comparison with our method to demonstrate its effectiveness and efficiency. These methods include the Human Baseline as well as some sophisticated agent-based automation approaches.

\textbf{Human Baseline} records the process of a human completing the instructions and is considered the golden solution for solving each task, as it reflects the best method based on human performance.

\textbf{AppAgent}~\cite{zhang2023appagent} introduces a framework with two phases: exploration and deployment. In the exploration phase, an agent learns app functions through self-learning or observation of humans, storing the knowledge in app-specific documents. During deployment, the agent uses these documents, along with the view hierarchy and screenshots, to plan and select actions. Each interactive element is labeled with bounding boxes and a unique index, improving the agent's accuracy in task execution.

\textbf{Mobile Agent(v2)}~\cite{wang2024mobile} is a multi-agent system for mobile device operation assistance, comprising planning, decision, and reflection agents. The system takes screenshots as input and utilizes additional modules such as the OCR model and qwen-vl-plus API, enabling more effective action generation in complex mobile operation tasks.

\textbf{Moba}~\cite{zhu2024moba} utilizes a two-level agent architecture with a Global Agent (GA) and a Local Agent (LA) to enhance mobile task automation. The GA interprets user commands and manages task planning, while the LA executes specific actions on the screen. The system takes as input both visual information and XML view hierarchy data to understand the mobile interface. For action execution, it employs a combination of OCR for text recognition and target localization to guide the selection of interactive elements.

\section{Evaluation Metrics}~\label{sec:metrics_details}
Before introducing the specific metrics for measuring the assistants, in order to better understand the subsequent calculations, we first define two sequences. The first is $\mathbf{a}_{\text{human}}^{q}=\{a_1,a_2,\dots,a_n\}$, representing the sequence of actions taken by a human to perform task $q$, and the corresponding $\mathbf{a}_{\text{agent}}^{q}=\{a_1,a_2,\dots,a_m\}$, representing the sequence of actions taken by the agent to perform task $q$. $n$ and $m$ represent the lengths of sequences $\mathbf{a}_{\text{human}}^{q}$ and $\mathbf{a}_{\text{agent}}^{q}$, respectively. Based on these sequences, we evaluate the performance of different methods from two key aspects: \textbf{Effectiveness} and \textbf{Efficiency}. Here, Effectiveness represents the success rate of the agent in completing tasks, while Efficiency reflects the speed or resource utilization during task execution.

\textbf{Effectiveness.} \textbf{Successful Rate (SR):} This metric measures the average proportion of successful task completions by the agent. A task is considered successful if the agent completes the instruction within 20 actions. \textbf{Completion Rate (CR)}~\cite{zhu2024moba}: Although many instructions may not be fully completed, the intermediate processes executed by the agent are also valuable for evaluation. The CR metric represents the proportion of correctly executed steps by the agent, relative to the total number of actions required to complete the task, using human operation as the ground truth. The formula for calculating the CR metric is:
$$
\mathbf{CR}=\frac{\sum_{q \in Q} \left| \mathbf{a}^q_{\text{human}} \cap \mathbf{a}^q_{\text{agent}} \right| }{\sum_{q \in Q}\left| \mathbf{a}^q_{\text{human}} \right|},
$$
where $Q$ is the set of instructions used to test the method. These two metrics measure the degree of task execution from the instruction and action levels, respectively.

\textbf{Efficiency.} In addition to task completion accuracy, the speed of task execution plays a crucial role in shaping the user experience in app scenarios. Therefore, we assess efficiency using three key metrics. First, it is essential to highlight the two primary time-consuming components of the agent: \textbf{(1) Subtask to Action:} The agent needs to map a task or subtask to an executable action sequence, which requires calling the action agent model. The number of times the action agent is called during this process is denoted as $\text{C}_{a}$. \textbf{(2) Executing Actions:} The agent must convert actions into executable commands, which involves using the grounding model or parsing actions. This time is represented by the length of the action sequence, $\left| \mathbf{a}_{agent} \right|$. Since AppAgent, Mobile Agent(v2), and Moba do not generate multiple actions at once, the $\text{C}_{a}$ value for these methods is equal to $\left| \mathbf{a}_{agent} \right|$. Next, we present three metrics to measure efficiency from different aspects. \textbf{Mapping Efficiency (ME)}, calculated as:
$$
    \mathbf{ME} = \frac{\sum_{q \in Q} \left| \mathbf{a}_{\text{human}}^{q} \right|}{\sum_{q \in Q} C_{a}}.
$$
This metric measures the efficiency of action sequence generation from the perspective of the action agent. A higher value indicates higher efficiency. Our method may generate multiple actions at once, leading to a $\mathbf{ME}$ greater than $1$. \textbf{Action Efficiency (AE)}, calculated as: 
$$
\mathbf{AE} = \frac{\sum_{q \in Q} \left| \mathbf{a}^{q}_{\text{human}} \right|}{\sum_{q \in Q} \left| \mathbf{a}^{q}_{\text{agent}} \right|}.
$$
This metric measures the efficiency of executing action sequences for different methods. A higher value indicates higher execution efficiency. \textbf{Average API Cost}, since in addition to plans and actions, other modules such as Memory and Reflection in different methods~\cite{zhu2024moba,wang2024mobile} may call the LLM API which is the primary consumer of time and computational resources. Therefore, we measure the overall execution efficiency of the architecture by the number of API calls required for the agent to generate each action in human actions $\mathbf{a}_{\text{human}}$, calculated as:
$$
    \mathbf{AAC} = \frac{\text{API}_{\text{count}}}{\sum_{q \in Q} \left| \mathbf{a}^q_{\text{human}} \cap \mathbf{a}^q_{\text{agent}} \right| }.
$$

\section{Subtask Case}~\label{sec:subtask_case}
In Table~\ref{tab:custome_subtask_example}, we present two examples, each containing a task and the corresponding subtasks decomposed by the plan agent in CHOP. As shown, our output not only includes basis subtasks but also features custom subtasks, highlighted in red. This demonstrates that our method can compensate for cases where the basis subtask cannot handle certain tasks by generating custom subtasks, thereby improving the quality of the generated subtasks.

In Table~\ref{fig:case_study}, we further present two examples showing that our basis subtasks can address both ineffectiveness and inefficiency issues. Specifically, in the first example, the task highlighted in blue is too complex to be executed by the downstream action agent. Our method breaks this blue subtask into two basis subtasks, making them simpler to execute, thus solving the ineffective subtask. Additionally, our method ensures more appropriate subtask granularity, such as using a single subtask for the sharing action, while without the restriction, two steps would be required. In the second example, the subtask highlighted in orange does not affect the task progression. Our method resolves this inefficiency by introducing a subtask in the critical path, thereby avoiding the inefficient subtask.

\section{Case Study}~\label{sec:case_study}
We present an example of the subtasks we executed in Figure~\ref{fig:case_study}. In this example, our method, due to the basis subtask, does not directly click ``Live'' on the homepage to find relevant streams. Instead, it uses the ``Search'' basis subtask to perform the search. Although this approach may involve more steps than directly navigating to the live page, it is more structured and reliable, reducing the chances of execution errors. Additionally, since the ``Search'' process is relatively fixed, we can have the action agent generate the entire action sequence for the search subtask in one call, reducing the number of action agent invocations.

\end{document}